\documentclass[a4paper]{article}

\usepackage{INTERSPEECH2022}
\usepackage[utf8]{inputenc}
\usepackage{lipsum}
\usepackage{multirow}
\usepackage{algorithm} 
\usepackage{algorithmicx}
\usepackage{algpseudocode}
\algdef{SE}[FUNCTION]{Function}{EndFunction}%
   [2]{\algorithmicfunction\ \textit{#1}\ifthenelse{\equal{#2}{}}{}{(#2)}}%
   {\algorithmicend\ \algorithmicfunction}

\newcommand\blfootnote[1]{%
  \begingroup
  \renewcommand\thefootnote{}\footnote{#1}%
  \addtocounter{footnote}{-1}%
  \endgroup
}

\title{Federated Pruning: Improving Neural Network Efficiency \\ with Federated Learning}
\name{Rongmei Lin$^1$, Yonghui Xiao$^2$, Tien-Ju Yang$^2$, Ding Zhao$^2$,  
\\ Li Xiong$^1$, Giovanni Motta$^2$, Françoise Beaufays$^2$}
\address{
  $^1$Emory University, $^2$Google LLC}
\email{\{rlin32,lxiong\}@emory.edu, \{yohu,tjy,dingzhao,giovannimotta,fsb\}@google.com}

\begin{document}

\maketitle
\blfootnote{This work was done while Rongmei Lin was an intern at Google.}

\begin{abstract}
Automatic Speech Recognition models require large amount of speech data for training, and the collection of such data often leads to privacy concerns. Federated learning has been widely used and is considered to be an effective decentralized technique by collaboratively learning a shared prediction model while keeping the data local on different clients devices. However, the limited computation and communication resources on clients devices present practical difficulties for large models. To overcome such challenges, we propose Federated Pruning to train a reduced model under the federated setting, while maintaining similar performance compared to the full model. Moreover, the vast amount of clients data can also be leveraged to improve the pruning results compared to centralized training. We explore different pruning schemes and provide empirical evidence of the effectiveness of our methods. 
\end{abstract}

\vspace{2mm}


\section{Introduction}
Neural network models have wide application in a variety of tasks, such as speech recognition, machine translation, and image recognition \cite{li2020comparison, dosovitskiy2020image,mccann2017learned}. The performance of trained models largely depends on the quality and the amount of training data. Federated learning (FL)~\cite{kairouz_2019_openproblem} provides a framework for leveraging the abundant data on edge devices with privacy preserved. However, FL faces several limitations in practice. One limitation is that the available memory on edge devices is highly limited. However, recent models are typically large, which makes on-device training challenging. For example, the successful model architecture for Automatic Speech Recognition (ASR), Conformer~\cite{gulati2020conformer}, has 130M parameters and requires 520MB memory solely for storing the parameters during training. Another limitation is that FL typically only updates the model parameters and leaves the model architecture unchanged. As a result, only model accuracy is improved but not model efficiency.

In this paper, we propose \emph{Federated Pruning (FP)} to address the limitations mentioned above. Because models are usually over-parameterized to facilitate training, there are many redundancies. Several methods have been explored to exploit such redundancies to improve model efficiency. Among them, pruning is one of the most successful methods and has been widely studied under centralized training settings. At a high level, it identifies and removes redundant parameters from an over-parameterized model. The proposed FP applies the same idea to improve efficiency of federated learning and also allows leveraging on-device data to potentially achieve better efficiency than centralized pruning.

The pruning method has been extensively studied in centralized fashion \cite{han2015deep,yang_2017_energy_pruning}. \cite{frankle2018lottery} shows that a well-initialized sub-network can match the accuracy of the full network and such sub-network was studed in centralized training \cite{han2015deep,yang_2017_energy_pruning}. The main issue of this approach is that the parameters deemed unimportant and pruned at an early iteration may turn out to be important at a later iteration. To address this problem, \cite{mocanu2018scalable, mostafa2019parameter} use different pruning method time-wise and model-wise.
Unlike these works focusing on centralized training, our work targets at federated learning and analyzes the impact of different pruning design decisions under this setting. A related work of model compression under the FL setting is Federated dropout \cite{guliani2021enabling}. Unlike our proposed method, federated dropout randomly generates reduced model and performs training on full model. Another preliminary work, PruneFL~\cite{jiang_2019_prunefl}, also applies pruning to federated learning. It adopts sparse pruning instead of structural pruning as used in this work, so the resultant model will be less efficient when running on devices in practice. Moreover, we evaluate the proposed FP with production-grade models and datasets, which better reflects the real condition of deployment.

\begin{figure}[t]
\centering
  \includegraphics[width=0.9\linewidth]{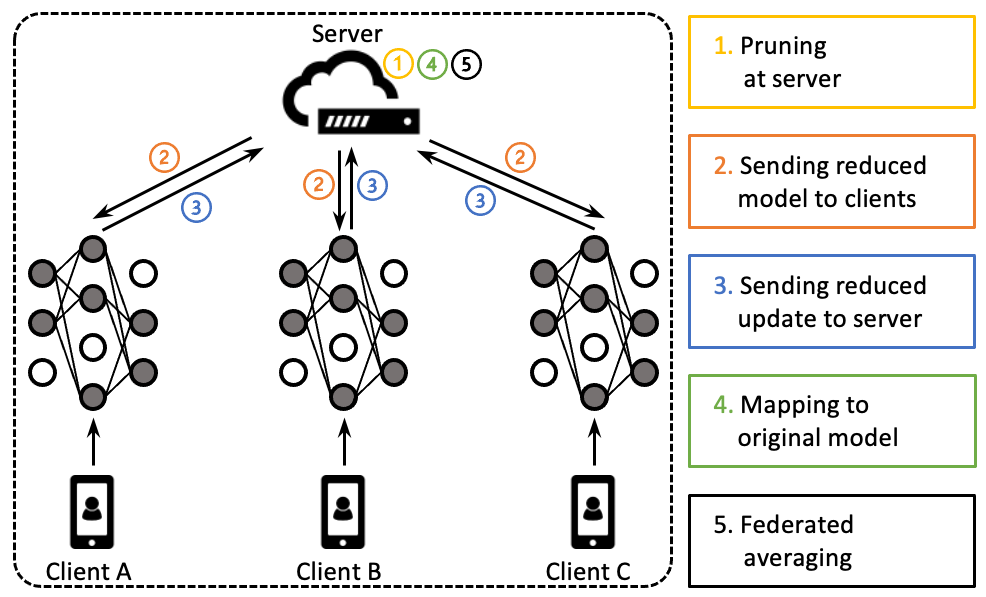}
  \caption{A federated round of the proposed Federated Pruning. The white circles denote removed parameters.}
  \label{fig:pipeline}
  \vspace{-6mm}
\end{figure}

In summary, this work has the following contributions:
\begin{itemize}
    \item \textbf{Improving the efficiency of federated learning:} We propose Federated Pruning (FP) to leverage on-device data to effectively prune redundant parameters from models. The resultant smaller models require less on-device memory to train and lower bandwidth for transporting models.
    \item \textbf{Exploring different pruning design decisions:} We explore and perform extensive ablation studies on two design decisions of pruning under federated learning: pruning patterns and pruning methods.
    \item \textbf{Proposing a novel approach for adaptive sparsity:} We propose a novel adaptive per-layer sparsity approach that dynamically allocates the target global sparsity level to each layer. Therefore, there is no need to manually select the per-layer sparsity levels.
    \item \textbf{Experimenting with production-grade environments: } We evaluate the proposed Federated Pruning with production-grade models and datasets, which better reflects the real condition of deployment.
\end{itemize}

\begin{algorithm}[t]
    \centering
    \footnotesize
    \caption{\footnotesize Federated pruning. Initialize the server model with $w^0$ and the binary pruning mask $M$ with \textit{ones} like ($w^0$). The $K$ clients are selected and indexed by $k$,  federated pruning rounds are indexed by $r$, and $n$ is the number of examples. \textit{Shrink}$(w, M)$ reduces the model size according to pruning mask $M$. \textit{Expand}$(w, M)$ maps the reduced model to original size. Related pruning methods include following functions: \textit{GetImportanceScore} ($w^{r}$), \textit{GenerateMask} ($w^{r}, r, S$). }\label{alg:federated_prune}
    \begin{algorithmic}[1]
    \State \textbf{Input:} Pre-trained dense ASR model: $w^0$
    \State \hskip 3.05em Binary pruning mask: $M$ of \textit{ones} like ($w^0$) 
    \State \hskip 3.05em Target sparsity level: $S$
    \State \hskip 3.05em FL rounds: $\Delta R, R^{fine-tune}, R^{end}$
    \State \textbf{Output:} Sparse ASR model: $w^{R^{end}}$
    \Function {FederatedPruning}{}
    \State initial sparsity level $s\gets 0$
    \For{each round $r = 0, 1, 2, ..., R^{fine-tune}-1$} 
    \If {$r\;mod\; \Delta R == 0$}  \Comment{Every $\Delta R$ rounds}
      \State \textit{GetImportanceScore} ($w^{r}$)
      \State $M$ = \textit{GenerateMask} ($w^{r}, r, s$)
      \If{$s<S$} 
      \Comment{Reaches refining phase if $s==S$}
      \State increase $s$
      \EndIf
    \EndIf
    \State $w^{r+1}$ = \textit{FPTrain($w^r, M$)}
    \EndFor
    \For{each round $r=R^{fine-tune}, ... , R^{end}$}
    \Comment{Fine-turning}
    \State{Reduce the server model with mask $M$}
    \State{Training the reduced model with standard FL.}
    \EndFor
    \State \Return $w^{R^{end}}$
    \EndFunction
    \Function {FPTrain}{$w^r, M$}
        \State $W^r \leftarrow$ \textit{Shrink}$(w^r, M)$ \Comment{Generate reduced model}
    \State Randomly select $K$ clients
    \State Server sends the reduced model $W^r$ to $K$ clients
        \For{each client $k$ \textbf{in parallel}}
        \State $\hat{W}_k^r\leftarrow$ \textit{ClientLocalUpdate}($k, W^r$)
        \State $\Delta W_k^r = W^r - \hat{W}_k^r$
        \State{Clients send $\Delta W_k^r$ to server}
        \EndFor
    \State $\Delta w_k^r \leftarrow$ \textit{Expand}$(\Delta W_k^r, M)$ \Comment{Map reduced updates}
    \State $\bar{w}^r = \sum_{k=1}^{K} \frac{n_k}{n}\Delta w_k^r$ \Comment{Federated Averaging \cite{mcmahan2016communication}}
    \State $w^{r+1}=w^r-\eta \bar{w}^r$
    \State \Return $w^{r+1}$
    \EndFunction
    \end{algorithmic}
\end{algorithm}

\section{Federated Pruning}
Figure~\ref{fig:pipeline} describes a federated round with the proposed Federated Pruning. The round starts from the generation of a set of variable masks on the server. The masks contain binary values signifying whether the parameters at the corresponding locations should be pruned (value 0) or not (value 1). The number of pruned values is determined by a given sparsity level, which is the ratio of pruned parameters. The server model is pruned based on the masks and sent to clients. Each client then trains this reduced model on its local data and returns the trained model. The server model finally aggregates the trained models from all the clients and moves on to the next federated round. Compared with standard federated learning, federated pruning prunes the model at the beginning of each round, and the pruned models are the ones trained on clients and transported.

\begin{figure}[t]
\centering
  \includegraphics[width=1\linewidth]{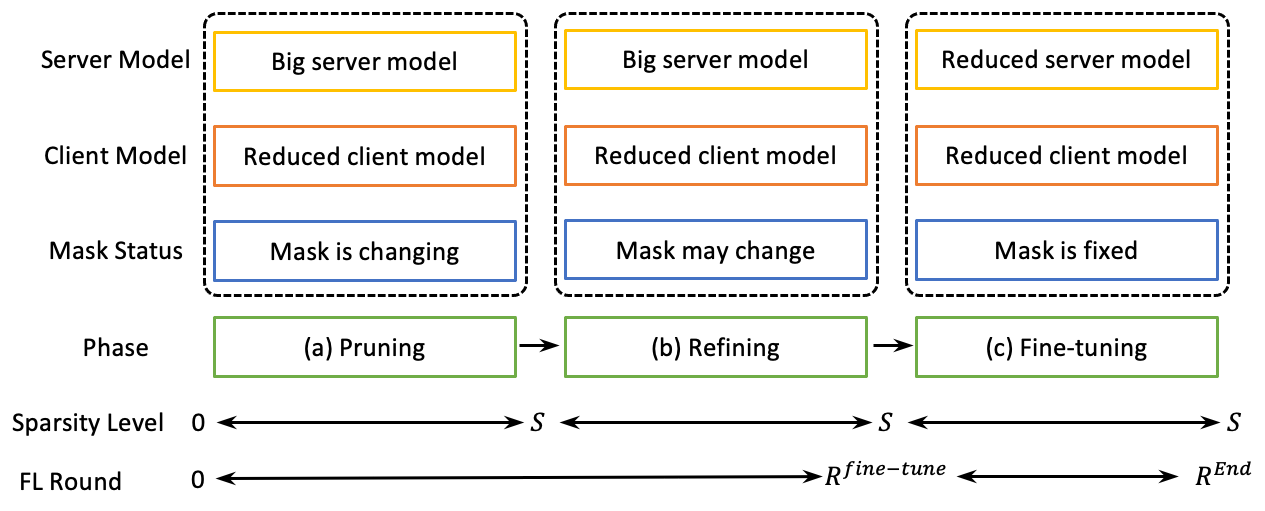}
  \caption{Three phases of Federated Pruning across FL rounds.}
  \label{fig:phase}
\end{figure}

Across federated rounds, we define three phases, (1) pruning, (2) refining and (3) fine-tuning, as shown in Figure~\ref{fig:phase}. In the first phase (pruning), the sparsity level ramps up from $0$ to the target sparsity level $S$. In the second phase (refining), the sparsity level is fixed at $S$, and the focus is refining the masks. Please note that there is a chance that a value in a mask flips from the $0$ to $1$, which means a pruned parameter revives. We show in Section~\ref{subsec:with_without_mask_refinement} that this phase is important for improving model quality. In the last phase (fine-tuning), the sparsity and the masks are fixed, and the focus is fine-tuning the remaining parameters. This phase ends when the pruned model converges.

We summarize the details of our method in Algorithm \ref{alg:federated_prune}. We consider one cloud server and $K$ edge devices (referred to as clients) with their local speech data. Let $w$ be the parameters of the full ASR model on the server side and $\{w_1, w_2, ... , w_k\}$ be the $K$ reduced models to be trained on clients. FP starts with pre-trained ASR model $w^0$ on the server side and the target sparsity level $S$. At the beginning of each $\Delta R$ rounds, the server computes the importance scores of all variables using \textit{GetImportanceScore()}. Then, a set of masks are generated by \textit{GenerateMask()} based on the importance scores and the current sparsity level $s$. In the generated masks, all variables with small importance scores will be removed by \textit{Shrink}$(w^r, M)$. The reduced model will be sent to clients for training, and mapped back to the full model when the server receives clients' updates by \textit{Expand($w, M$)}. When mapping the reduced model back, the masked regions (which are not sent to clients) will have 0 aggregated updates. Then the standard federated averaging process will be used to aggregate all clients updates. When the sparsity level $s$ reaches the target level $S$, it enters the second phase of mask refinement. When the round $r$ is equal to $R^{fine-tune}$, FP moves on to the last phase. We reduce the server model with the mask $M$, so that the server model is the same as client models. Finally, the reduced model will be trained until convergence. %

\subsection{Pruning patterns}
We discuss the structure of the pruned variables, i.e. removed variables, as pruning patterns. Pruning patterns can be broadly categorized as: unstructured pruning \cite{dynamic-sparsity-nonstructured} that prunes the less salient connections over any nodes, and structured pruning \cite{slimmable-nn} that prunes on larger structures such as channel or layer. Since our objective is to reduce the actual model size, the training memory on clients' devices and the communication between server and clients, we use structured pruning to remove the \emph{slices} of the variables, and thus physically reduce the model size. We implemented following pruning patterns:

\begin{itemize}
    \item Whole row / column: prune the entire row or column of the two-dimensional weight matrices $W$.
    \item Half row / column: evenly partition the two-dimensional variables $W$ into $[W_1,W_2]$ and prune each half of the row or column.
\end{itemize}
For higher (larger than 2) dimensional matrices, we first reshape them to two dimensional space, apply the above patterns, and transform back to the original dimensional space.

\subsection{Pruning methods} 
\label{sec:pruning_methods}
We discuss how to decide the salience, i.e. the importance score, of variables. We use three methods including weight magnitude of variables, momentum of gradient magnitude of the variables and the multiplication of weight and gradient magnitude to measure the salience. At every $\Delta R$ round, we compute the $l1$ and $l2$ norm of the three methods as importance scores. Then given the sparsity level $s$, a threshold of the importance score can be derived by sorting the scores. The regions with less importance scores (i.e. smaller than the threshold) will be removed from the clients' model to form the reduced model. On server side, the regions with less importance scores are still kept in the pruning and refining phases, and will be removed in the final fine-tuning phase.

To achieve the target sparsity level $S$, we use pruning schedule to denote such process in the pruning phase. In this work, we explore the following two schedules:
\begin{itemize}
    \item Constant sparsity level: instantly prune to the target sparsity level $S$ at the beginning;
    \item Step-based sparsity level: gradually increase the sparsity level w.r.t current round $r$.
\end{itemize}

\subsection{Mask refinement} \label{sec:mask_refine}
In the refining phase, the masks are still re-generated. Because the masked regions in the model get 0 updates, their importance scores remain the same. For unmasked regions, their variables as well as the importance scores are updated. If the importance scores of some unmasked regions get smaller, then they might be replaced by the masked variables according to the rank of importance scores. Therefore, the masks are refined and re-trained in this phase. In section 4, we also explore the variants of with and without mask refinement to show its utility. Note that the gradient based importance scores will yield relatively stale pruning mask, the masked regions will not have gradients and lose the ability of regrowth.

\subsection{Adaptive per-layer sparsity}
The layers of a deep neural network have different impacts on model accuracy. Based on the observations in \cite{zhou2021exploring}, layers can be categorized as either “ambient” or “critical”. Take the Conformer model \cite{gulati2020conformer} as example, “ambient” layers have little impact whereas pruning the “critical” layers will lead to severe quality degradation. Hence the sparsity level should get customized based on the importance of each layer. The sparsity level in the above federated pruning is unified among layers. Adaptive per-layer sparsity is proposed to allow dynamic sparsity level reallocation among layers.

\begin{figure}[t]
\centering
  \includegraphics[width=1\linewidth]{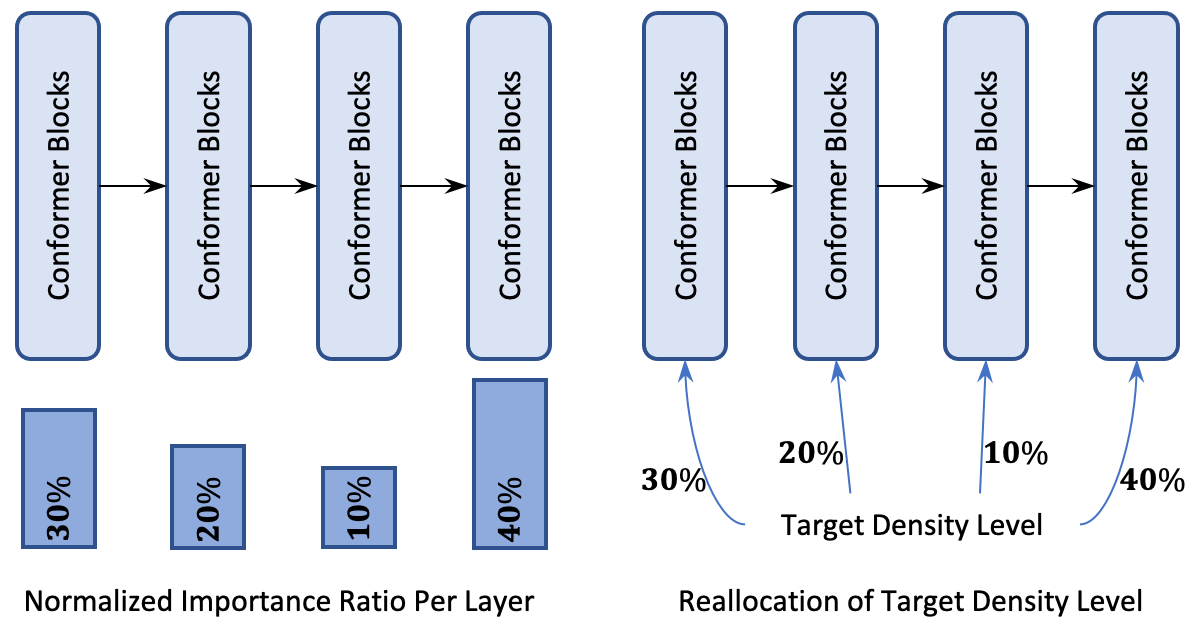}
  \caption{Reallocation process}
  \label{fig:reallocate}
\end{figure}

We utilize a heuristic agent to determine the layer-wise sparsity level in two steps. First, we use predefined rules to measure the importance of each layer, such as the mean momentum of model deltas per layer or averaged weight magnitude per layer. Second, we assign the weighted sparsity level to each layer using estimated importance score. The less important layers get larger sparsity level. We use the averaged weight magnitude $||w_i||$ per layer as the importance score in this paper. As shown in Figure \ref{fig:reallocate}, the layers with larger magnitude are more important than those with smaller magnitude, thus the layer-wise density levels (1 - sparsity level) are assigned using the following equation for the $L$ layers in the model.
\begin{equation}
    LayerDensity = \frac{(1-TargetSparsity)*||w_i||}{\sum_{i=1}^{L}||w_i||}
\end{equation}

\section{Experiment Results}
\subsection{Experiment settings}
\noindent \textbf{Model architecture.}
We implemented the Federated Pruning in a distributed learning simulator. We use the state-of-the-art ASR model Conformer-transducer \cite{li2021better} as our base model. The model consists of 17 512-dimensional conformer encoder layers, a 640-dimensional embedding prediction network and a 2048-dimensional fully-connected joint network. In order to fit in our federated setting \cite{hsieh2020non, reddi2020adaptive}, as suggested in \cite{guliani2021enabling}, we change the original batch normalization to group normalization \cite{wu2018group}.
All the experiments share a same baseline model as initialization $w^0$. We follow the same settings of server/clients optimizer, SpecAugment \cite{park2019specaugment} and number of clients as in previous work \cite{guliani2021training}. The baseline model is trained from scratch for 30k federated rounds. Our metric Word Error Rate (WER) on the baseline model is shown in Table \ref{tab:adaptive}.

\noindent \textbf{Dataset.} We train and evaluate the proposed model on the public LibriSpeech \cite{panayotov2015librispeech} corpus, which consists of 970 hours of labeled speech. We also explore the performance on industry-scale data collected from different domains as described in \cite{narayanan2019recognizing}. These multi-domain utterances contain 400k hours of speech and span domains of search, farfield, telephony and YouTube. Our work abides by Google AI Principles \cite{google}, all datasets are anonymized and hand-transcribed.

\subsection{Federated pruning results}

\begin{figure}[t]
\centering
  \includegraphics[width=1\linewidth]{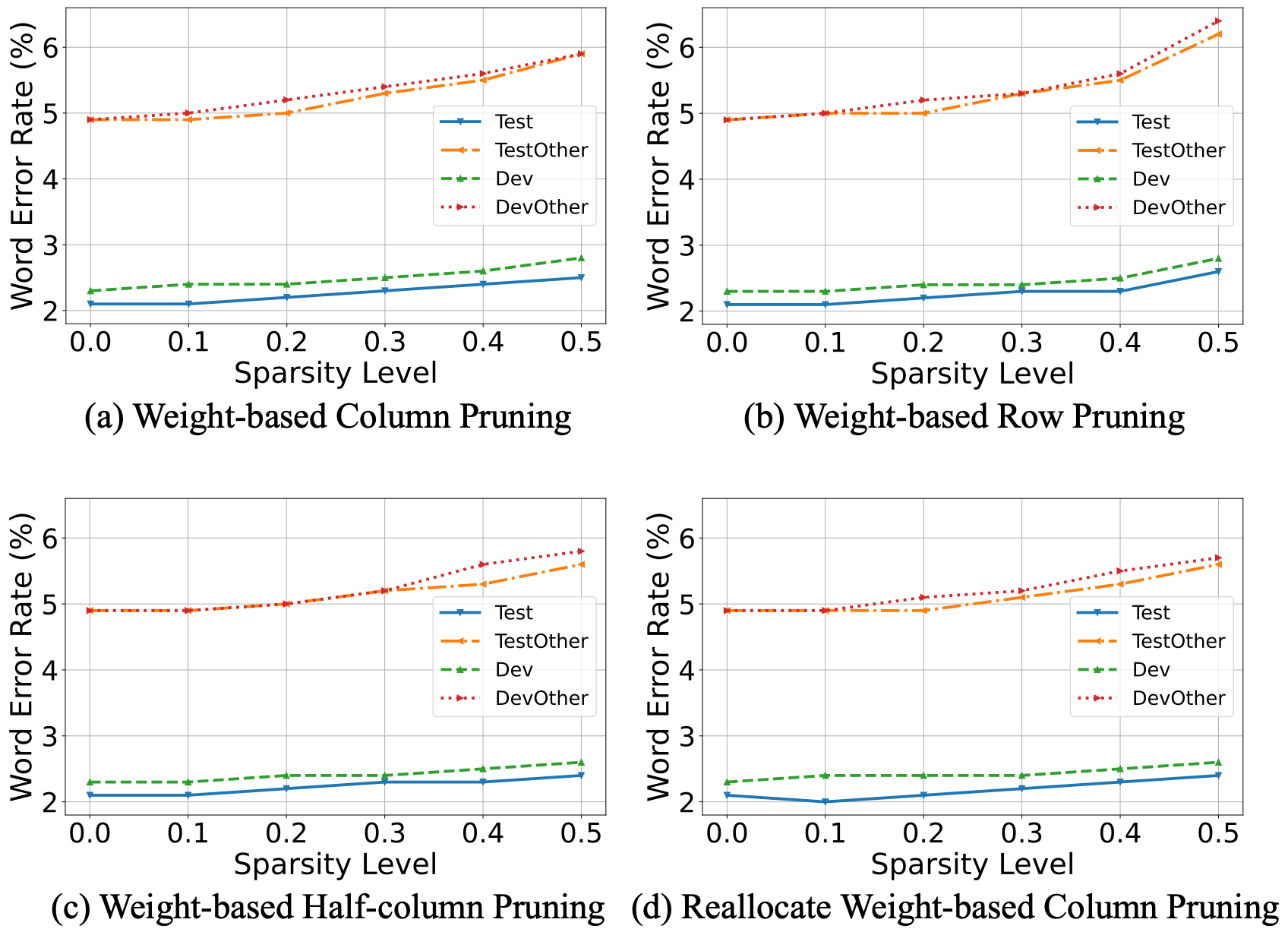}
  \caption{Experiment results of different settings of Federated Pruning on the Librispeech datasets.}
  \label{fig:wer_curve}
\end{figure}

We prune all variables in each layer except the 1-D vector variable and the convolution layer which has specific utility and few parameters. The target sparsity level is the unified pruning percentage assigned for each variable. If we adopt the whole block pruning that prunes the entire column on $W$, it will further zero-out the corresponding row in next variable $W'$ within the feed forward module in the Conformer blocks. The actual zero-parameter ratios of to-be-pruned variables are slightly higher than the target sparsity level. The detailed numbers are listed in Table \ref{tab:ratio}. 

Figure \ref{fig:wer_curve} shows a rise in WERs on 4 Librispeech data subsets with the increase of sparsity levels. We also observed obvious quality degradation when the sparsity level $>$ 30\% in all pruning schemes. Concretely, 0.2\% absolute increase on WER was observed on both Dev and Test evaluation set in our default setting Weight-based column pruning when the sparsity level reached 0.3. As we mentioned in Section \ref{sec:pruning_methods}, different \emph{pruning methods} can be used to estimate the importance of variables. We conduct ablation experiments of different measurements on the Librispeech dataset. Our empirical finding in Table \ref{tab:wer_methods} suggests the weight-based score achieves similar WER as other metrics, while it is also the most communication efficient and stable metric. Thus, we rely on the weight-based score as the importance metric. In terms of the \emph{pruning patterns}, we compare the WERs of several FP variants at sparsity level 50\% as illustrated in Figure \ref{fig:wer_curve}. One can observe that column pruning and especially the half-column pruning consistently outperforms the row-based pruning. Thus, all experiments below adopt weight magnitude as pruning method and whole column as pruning pattern. With the subsequent fine-tuning phase, the pruning schedule has very limited effect on the WER, so we further fix the constant sparsity level as the default setting.

\begin{table}[h]
  \caption{Zero-Parameter Ratio w.r.t Target Sparsity Level}
  \label{tab:ratio}
  \centering
  \scalebox{1.0}{
  \begin{tabular}{c|ccccc}
    \toprule
    Sparsity Level & 0.10 & 0.20 & 0.30 & 0.40 & 0.50 \\
    \midrule
    Zero-Param Ratio & 0.136 & 0.264 & 0.384 & 0.496 & 0.60 \\
  \bottomrule
\end{tabular}
}
\end{table}

\begin{table}[h]
  \caption{WERs of different pruning methods}
  \label{tab:wer_methods}
  \centering
  \scalebox{1.0}{
  \begin{tabular}{c|ccccc}
    \toprule
    \multirow{2}{*}{\textbf{Pruning Methods}}& \multicolumn{5}{c}{\textbf{WER on Test with sparsity level}} \\
                            & \textit{0.10} & \textit{0.20} & \textit{0.30} & \textit{0.40} & \textit{0.50} \\
    \midrule
    weight based & 2.1 & 2.2 & 2.3 & 2.4 & 2.5 \\
    gradient based & 2.2 & 2.3 & 2.3 & 2.3 & 2.4 \\
    weight$\times$gradient based & 2.1 & 2.2 & 2.2 & 2.3 & 2.4 \\
  \bottomrule
\end{tabular}
}
\end{table}

\subsection{Adaptive per-layer sparsity results} We propose the adaptive per-layer sparsity to make our method more flexible and dynamic, which allows reallocation of target pruning budgets among layers. Table \ref{tab:adaptive} demonstrates that, compared to federated pruning with unified sparsity, the adaptive sparsity achieves lower WERs on all evaluation sets with 50\% sparsity level.

\begin{table}[h]
  \caption{WERs of unified / adaptive sparsity at Sparsity 50\%.}
  \label{tab:adaptive}
  \centering
  \scalebox{1.0}{
  \begin{tabular}{c|cccc}
    \toprule
    \multirow{2}{*}{\textbf{Exp.}}& \multicolumn{4}{c}{\textbf{WER}} \\
                                    & \textit{Test} & \textit{TestOther} & \textit{Dev} & \textit{DevOther} \\
    \midrule
    Baseline & 2.1 & 4.9 & 2.3 & 4.9 \\
    Unified Sparsity & 2.5 & 5.9 & 2.8 & 5.9 \\
    Adaptive Sparsity & 2.4 & 5.6 & 2.6 & 5.7 \\
  \bottomrule
\end{tabular}
}
\vspace{-5mm}
\end{table}

\subsection{With and without mask refinement} 
\label{subsec:with_without_mask_refinement}
To actually reduce the model size, 
the masked regions on server model are eventually zero-out by FP. On the other hand, the \emph{Mask Refinement} phase introduced in Section \ref{sec:mask_refine} maintains the original values for masked regions. The pruned variables are allowed to grow back, leading to higher flexibility and thus, lower WERs as shown in Table \ref{tab:late}.

\begin{table}[h]
  \caption{WERs of w/o and w/ Mask Refinement at Sparsity 40\%.}
  \label{tab:late}
  \centering
  \scalebox{1.0}{
  \begin{tabular}{c|cccc}
    \toprule
    \multirow{2}{*}{\textbf{Exp.}}& \multicolumn{4}{c}{\textbf{WER}} \\
                                    & \textit{Test} & \textit{TestOther} & \textit{Dev} & \textit{DevOther} \\
    \midrule
    w/o Mask Refine & 2.3 & 5.5 & 2.6 & 5.7 \\
    w/ Mask Refine & 2.3 & 5.3 & 2.5 & 5.3 \\
  \bottomrule
\end{tabular}
}
\vspace{-5mm}
\end{table}

\subsection{Results on short-form multi-domain dataset} Finally, we show the results on the large-scale short-form multi-domain dataset. The reduced model is trained on our multi-domain utterances and evaluated on the short-form dataset. Table \ref{tab:vs} demonstrates the WERs on different sparsity levels. We conclude that our model can still achieve comparable performance to the baseline model (with sparsity level $0.0$) on challenging dataset in the low sparsity level setting.

\begin{table}[h]
  \caption{WERs of federated pruning on the voice search dataset.}
  \label{tab:vs}
  \centering
  \scalebox{1.0}{
  \begin{tabular}{c|cccccc}
    \toprule
    Sparsity Level & 0.0 & 0.10 & 0.20 & 0.30 & 0.40 & 0.50 \\
    \midrule
    WER & 6.4 &  6.7 & 7.0 & 7.4 & 7.9 & 8.9 \\
  \bottomrule
\end{tabular}
}
\vspace{-2mm}
\end{table}

\section{Conclusion}
We proposed the Federated Pruning method to find the efficient reduced model in FL settings. There are two advantages of FP. First, the training cost of clients, including the on-device training memory and communication, are alleviated due to the pruned and reduced model. Second, the model pruning quality can also be improved with the vast amount of clients data. We also proposed a new pruning method of the layer-wise sparsity level reallocation to improve the pruning quality. We showed in experiments that the FP trained model can achieve comparable quality of the full model. 

\clearpage
{
\bibliographystyle{IEEEtran}

}

\end{document}